\documentclass[conference]{IEEEtran}
\IEEEoverridecommandlockouts
\usepackage{cite}
\usepackage{amsmath,amssymb,amsfonts}
\usepackage{graphicx}
\usepackage{subcaption}
\usepackage{siunitx}
\usepackage{textcomp}
\usepackage{xcolor}
\def\BibTeX{{\rm B\kern-.05em{\sc i\kern-.025em b}\kern-.08em
    T\kern-.1667em\lower.7ex\hbox{E}\kern-.125emX}}

\usepackage{booktabs}
\usepackage{comment}    
\usepackage{amsmath,amssymb,amsfonts} 
\usepackage{xcolor} 

\usepackage[hidelinks]{hyperref}

\usepackage[noend]{algpseudocode}
\usepackage{algorithm}

\definecolor{cerulean}{rgb}{0.0, 0.48, 0.65}
\algdef{SE}[DOWHILE]{Do}{doWhile}{\algorithmicdo}[1]{\algorithmicwhile\ #1}
\newcommand{\lam}{\gets}
\renewcommand{\Comment}[1] {\hfill\textit{\textcolor{cerulean}{$\triangleright$~#1}}} 
\makeatletter
\algnewcommand{\OnlyComment}[1]{\hskip\ALG@thistlm\textit{\textcolor{cerulean}{$\triangleright$~#1}}}
\algnewcommand{\LineIndentComment}[1] {\Statex \hskip\ALG@thistlm\hskip\algorithmicindent\textit{\textcolor{cerulean}{$\triangleright$~#1}}} 
\algnewcommand{\LineComment}[1] {\Statex \hskip\ALG@thistlm\textit{\textcolor{cerulean}{$\triangleright$~#1}}} 
\makeatother

\newcommand{\thesys}[1]{TrimTuner}
\newcommand{\ts}[1]{\thesys{}}

\newcommand{\pvs}{\vspace{-10pt}}
\newcommand{\mynewpar}[1]{\pvs~\\\noindent{\bf #1}}

\makeatletter
\newcommand{\shorteq}{%
  \settowidth{\@tempdima}{-}
  \resizebox{2mm}{\height}{=}%
}
\makeatother
    
\begin{document}

\title{\ts{}: Efficient  Optimization of Machine Learning Jobs in the Cloud via Sub-Sampling
\thanks{This research was partially supported by FCT (POCI-01-0247-FEDER-045915 and UIDB/50021/2020) and NSA (Award No. H9823018D0008). The extensive evaluation presented was made possible due to the grant received from the AWS Cloud Credits for Research Program.}
}

\author{\IEEEauthorblockN{Pedro Mendes$^{1}$, Maria Casimiro$^{1,2}$, Paolo Romano$^{1}$, David Garlan$^{2}$}
\IEEEauthorblockA{
\textit{$^{1}$INESC-ID and Instituto Superior Técnico, Universidade de Lisboa}\\
\textit{$^{2}$Institute for Software Research, Carnegie Mellon University}
}
}

\maketitle

\begin{abstract}

This work introduces \ts{}, the first system for optimizing machine learning jobs in the cloud to exploit sub-sampling techniques to reduce the cost of the optimization process, while keeping into account user-specified constraints. \ts{} jointly optimizes the cloud and application-specific parameters and, unlike state of the art works for cloud optimization, eschews the need to train the model with the full training set every time a new configuration is sampled. Indeed, by leveraging sub-sampling techniques and data-sets that are up to 60$\times$ smaller than the original one, we show that \ts{} can reduce the cost of the optimization process by up to 50$\times$. 

Further, \ts{} speeds-up the recommendation process by 65$\times$ with respect to state of the art techniques for hyper-parameter optimization that use sub-sampling techniques. The reasons for this improvement are twofold: i) a novel domain specific heuristic that reduces the number of configurations for which the acquisition function has to be evaluated;
ii) the adoption of an ensemble of decision trees
that enables boosting the speed of the recommendation process by one additional order of magnitude.


\end{abstract}


\section{Introduction}


Training machine learning (ML) jobs on the cloud represents the \textit{de facto} standard approach today for larges models. Existing ML models are in fact increasingly complex~\cite{aml19} and their training procedure sometimes involves an enormous amount of computational resources, which the cloud can provision in an elastic and on-demand fashion, sparing users and enterprises from colossal upfront capital investments. Further, by taking advantage of economies of scale, cloud providers can drastically reduce their internal operational costs. This ultimately reflects into cost savings for the end users. 

Nonetheless, a key problem that users face is that modern cloud providers offer a large spectrum of heterogeneous virtual machine (VM) types, optimized for different types of resources and with different costs. For instance, at the time of writing, Amazon Web Services (AWS) EC2 offers approximately 300 different VM types~\cite{aws:flavors}. The problem is further exacerbated by the fact that the (distributed) training process of ML jobs exposes several hyper-parameters --- such as the batch size considered in each training iteration or the frequency of synchronization among workers. The optimal configuration for these parameters can be substantially affected by the choice of the type and number of provisioned cloud resources~\cite{lynceus}.

As such, end users who wish to train their ML models in cloud infrastructures are  faced with  a complex constrained optimization problem:  determining which type/amount of cloud resources and model hyper-parameters to use in order to maximize the model's accuracy while enforcing constraints on the maximum cost and/or time of the training process.

Given the complexity of modelling the dynamics of modern ML and cloud platforms via white-box methods, a common approach in the literature is to rely on black-box modelling and Bayesian Optimization (BO) techniques~\cite{cherrypick,lynceus,arrow}. These techniques have the key advantage of requiring no prior knowledge of the target ML model to be optimized, and as  such require the target ML model to be deployed and trained in several (cloud/hyper-parameter) configurations. The corresponding measurements of accuracy and execution cost/time are then used to build black-box models, e.g., Gaussian Processes (GPs). These models  guide the optimization process by recommending which configurations to test next, balancing exploration (of unknown regions of the configuration space) and exploitation (of the models' current knowledge)  via different model-driven heuristics, a.k.a., acquisition functions~\cite{Practical_BO}.

Unfortunately, these systems suffer from a severe limitation: each time a configuration is tested, the target ML model needs to be trained on its entire data-set.
As such, the  optimization process  can  become prohibitively  expensive (and slow) if the target model is meant to be trained on massive data-sets, as it is increasingly the case in practice~\cite{aml19}.

In this work, we tackle this problem  by presenting \ts{}, the first system for optimizing the  training of ML jobs in the cloud that exploits data sub-sampling techniques to enhance the efficiency of the optimization process. 

\ts{} considers the problem of identifying the cloud and hyper-parameter configuration that maximizes the accuracy of a ML model, while ensuring that user-defined constraints on the efficiency (e.g., cost or execution time) of the training process are complied with.

\ts{} deploys the target job using sub-sampled data-sets (up to 60$\times$ smaller than  the original one in our experiments) and constructs predictive models that keep into account how shifts of the data-set size affect both the quality (i.e., accuracy) and training efficiency (i.e., cost or execution time) of the target model. These models are then used within a novel acquisition function that aims at estimating the advantage of testing a new configuration \textbf{x} using a data-set that is $s\times$ smaller than the original one ($s\in$[0,1]) by weighing in two  main factors: (i) the expected information gain~\cite{informationgain} that testing $\langle \textbf{x}, s \rangle$ will yield on the configuration that maximizes accuracy when using the full data-set ($s$=1), and  (ii)  the likelihood that the latter configuration $\langle \textbf{x}, $s$=1 \rangle$ meets the user-specified constraints.

Overall, this paper makes  contributions of both a methodological and practical nature.
From a methodological perspective, \ts{} builds on recent systems for hyper-parameter tuning of ML models~\cite{fabolas,hyperband,bohb}, which have first investigated the use of sub-sampling techniques, and extends them in a number of ways: (i) by supporting the enforcement of additional user-specified constraints;  (ii) by jointly optimizing the model's hyper-parameters and the selection of the cloud configuration (number and type of virtual machines); (iii) by introducing a novel domain-specific heuristic,  named 
``Constrained  Expected Accuracy'', that accelerates the recommendation process by a factor of up to 2$\times$ when compared to state of the art approaches~\cite{cmaes,direct}; (iv) by proposing the adoption of an ensemble~\cite{bagging} of Decision Trees (DTs) \cite{decisionTrees} as a light-weight alternative to Gaussian Processes (the \textit{de facto} standard modelling approach for BO), which enables boosting the speed of the recommendation process by one additional order of magnitude.

From a practical perspective, when compared to state of the art BO-based systems for the (constrained) optimization of ML jobs, such as Lynceus~\cite{lynceus} or Cherrypick~\cite{cherrypick}, \ts{} reduces the cost  of the  exploration process by up to 50$\times$ thanks to the use of sub-sampling techniques.  Another practical  contribution of this work is making  available to the community the data-sets obtained for the evaluation of \ts{}, which consider the training of three neural networks via TensorFlow on AWS EC2  over a search space (encompassing both model hyper-parameters and cloud-related parameters)  composed of approximately 1400 configurations, whose collection incurred a cost of approximately $\$$1200 and took about two months.

The remainder of this paper is structured as follows: in \S~\ref{sec:rw} we describe related work; \S~\ref{sec:trimtuner} presents \ts{}; \S~\ref{sec:eval} evaluates our contributions and finally \S~\ref{sec:conclusion} concludes the paper.

\section{Related Work}
\label{sec:rw}

This section discusses related works in the areas of cloud optimization and Bayesian optimization.

\mynewpar{Cloud Optimization Approaches.}
Recent works to find the optimal configuration to deploy jobs in the cloud, such as CherryPick~\cite{cherrypick}, Ernest~\cite{ernest}, Paris~\cite{paris} and Arrow~\cite{arrow}, focus solely on optimizing cloud related parameters and disregard the possibility of learning from related tasks. Lynceus~\cite{lynceus} was the first work to highlight the relevance of jointly optimizing the cloud configuration and the 
hyper-parameters affecting the distributed training process of ML models. 

The key difference between \ts{} and these systems is its reliance on sub-sampling techniques to reduce the cost of testing configurations during the optimization process. As we discuss in \S~\ref{sec:eval}, this allows \ts{} to reduce the cost and duration of the optimization process by up to a factor of 50 and 65, respectively. 

Another key aspect of \ts{}, which differentiates it from systems like Paris~\cite{paris}, Quasar~\cite{quasar}, or Arrow~\cite{arrow} is that it does not rely on \textit{a priori} knowledge of similar types of jobs --- whose representativeness constitutes a key assumption on which the accuracy of the optimization process hinges. Conversely, \ts{} (analogously, e.g., to CherryPick and Lynceus) operates in a fully on-line fashion.


\mynewpar{Bayesian Optimization.} \ts{}, similarly to other recent systems for the optimization of cloud jobs~\cite{cherrypick,lynceus}, relies on a generic optimization methodology, known as (model-based) Bayesian Optimization (BO). BO has been adopted in a wide range of application domains including self-tuning of transactional memory systems~\cite{proteustm,zengIPDPS18}, databases~\cite{ituned} and of hyper-parameters of ML models~\cite{fabolas,bohb,mtbo}.

BO aims to identify the optimum $\textbf{x}^*$ of an unknown function  $f:\mathbb{X} \rightarrow \mathbb{R}$ and operates as follows~\cite{boTutorial}: (i) $f$ is evaluated (i.e., tested or sampled) over $N$ initial configurations, $\textbf{x}_i$, selected at random so as to build an initial training set $\mathcal{S}$ composed of pairs $\langle \textbf{x}_{i},f(\textbf{x}_{i})\rangle$; (ii) $\mathcal{S}$ is used  to train a black-box model (typically a Gaussian Process~\cite{gaussianprocesses}) that serves as a predictor/estimator of the unknown function $f$; (iii) an \textit{acquisition function}, noted $\alpha$, is used to exploit the model's knowledge (and related uncertainty) to determine which configuration to evaluate next by balancing exploitation of model's knowledge --- recommending configurations that the model deems to be optimal --- and exploratory behaviours --- recommending configurations whose knowledge can reduce the model's uncertainty and enhance its accuracy; (iv) the process is iteratively  repeated until a stopping condition is met, e.g., after a fixed budget is consumed or if the gains from further sampling are predicted to be marginal by the model (e.g., below a fixed threshold).

The definition of the acquisition function is arguably one of the most crucial components of BO methods and in the literature there exists a number of proposals. Expected Improvement (EI) is probably the most well-known acquisition function. As the name suggests, EI (Eq.~\eqref{eq:ei}) uses the probability distribution of observing a value $f(\textbf{x})$ at $\textbf{x}$, predicted by the model trained on data-set $\mathcal{S}$, in order to measure the expected amount by which evaluating $f$ at 
$\textbf{x}$ can improve over 
the current best value or 
\textit{incumbent} 
$\eta$:

\begin{equation}
\alpha_{EI}(\textbf{x})=\int max(0,f(\textbf{x})-\eta)p(f(\textbf{x})|\mathcal{S})df(\textbf{x})
\label{eq:ei}
\end{equation}

Entropy Search (ES)~\cite{entropySearch} is an alternative acquisition function that chooses which configurations to evaluate by predicting the corresponding information gain on the optimum, rather than aiming to evaluate near the optimum (as in EI). ES (Eq.~\eqref{eq:es}) is based on the probability distribution
$p_{opt}(\textbf{x} |  \mathcal{S})$, namely
the likelihood that a configuration $\textbf{x}$ belongs to the set of optimal configurations for $f$, given the current observations in $\mathcal{S}$. 
The information gain deriving from testing $\textbf{x}$ is computed using the expected Kullback-Leibler divergence (relative entropy)
between $p_{opt}(\cdot | \mathcal{S} \cup \{\textbf{x}, y\})$ and the uniform distribution
$u(\textbf{x})$, with expectations taken over the model-predicted probability of obtaining  measurement $y$ at $\textbf{x}$:

\vspace{-2mm}
\begin{equation}
\begin{aligned}
    \alpha_{ES}(\textbf{x}) = &~\mathbb{E}_{p(y|\textbf{x},\mathcal{S})} \left[ \int p_{opt}(\textbf{x'}|\mathcal{S} \cup \{\textbf{x},y\})  \right. \\ 
    &  \left. \cdot \log \frac{p_{opt}(\textbf{x'}|\mathcal{S} \cup \{\textbf{x},y\}) }{u(\textbf{x'})} d\textbf{x'} \right] 
    \label{eq:es}
\end{aligned}
\end{equation}

Despite being numerically much more complex to compute than EI, ES allows for quantifying to what extent testing a configuration $\textbf{x}$ will give the model knowledge about the optimum $\textbf{x}^*$, where generally $\textbf{x}\neq \textbf{x}^*$. In the context of hyper-parameter tuning of ML models, this property of ES has been exploited by MTBO~\cite{mtbo} and FABOLAS~\cite{fabolas} to trade off the information gain and the cost (i.e., execution time) of training a ML model in a configuration $\langle \textbf{x},s \rangle$, where $\textbf{x}$ denotes a hyper-parameter's configuration and $s\in[0,1]$ the sub-sampling rate applied to the original/full model's training data-set.

More precisely, in FABOLAS, the acquisition function (Eq.~\eqref{AF-F}) for $\langle \textbf{x},s \rangle$ is defined as the ratio between the information gain on the configuration that maximizes accuracy for the full data-set ($s$=1) and the (predicted) cost of training the model with a sub-sampling  rate  $s$ and hyper-parameters $\textbf{x}$:
\begin{equation}
\begin{aligned}
    \small
    \alpha_{F}(\textbf{x},s) 
    = & \mathbb{E}_{p(y|\textbf{x},s,\mathcal{S})} \left[ \int p_{opt}^{s=1}(\textbf{x'}|\mathcal{S} \cup \{\textbf{x},s,y\})  \right. \\ 
    &  \left. \log \frac{p_{opt}^{s=1}(\textbf{x'}|\mathcal{S} \cup \{\textbf{x},s,y\}) }{u(\textbf{x'})} d\textbf{x'} \right] \cdot \frac{1}{C(\textbf{x},s)} 
    \label{AF-F}
\end{aligned}
\end{equation}
\ts{} builds on these approaches and extends them in several ways. 
First, \ts{} supports the definition of additional (independent) constraints, e.g., on cloud  cost and/or execution times of  training/querying the model. This is achieved by extending the acquisition function to factor in the probability that the new incumbent will comply with the constraints, which will be discovered after updating the model with the observation of a (possibly sub-sampled) configuration $\langle \textbf{x},s \rangle$. Unlike existing constrained versions of ES-based acquisition functions, such as Predictive Entropy Search with Constraints (PESC)~\cite{pesc} and constrained Max-value Entropy Search (cMES)~\cite{cmes}, the proposed acquisition function does not make use of Bochner’s theorem for a spectral approximation, which allows \ts{} to use GPs with non-stationary kernels (as in FABOLAS) or lightweight (ensembles of) decision trees.
Further, \ts{}  jointly  optimizes  the    hyper-parameters of the training process and  the   cloud  configuration --- which, as already mentioned, is crucial to maximize the cost efficiency of the recommended configuration~\cite{lynceus}.

Finally, the numerical computation of the ES (and  of any acquisition function based on ES, like FABOLAS' and \ts{}'s) is onerous. \ts{} introduces two mechanisms to accelerate the recommendation process: (i) a novel domain-specific heuristic, that is used to estimate which configurations 
are most likely to yield the highest values of the acquisition function --- thus restricting the number of configurations for which the acquisition function is evaluated; (ii) differently from FABOLAS and MBTO, which rely on GPs to estimate the probability distribution of the accuracy and cost of unknown configurations, \ts{} relies on an ensemble of decision trees that, as we will show in \S~\ref{sec:eval}, achieves comparable accuracy while enabling speed-ups of up to 14$\times$.

\begin{algorithm}[t]
    \caption{\ts{}'s pseudo-code}
    \footnotesize
    \label{alg:trimtuner}
    \begin{algorithmic}[1]
    
    \Function{\ts{}}{$M$, $\mathbb{X}$, $\textbf{Q}$} 
    \LineComment{$M$: Model to be trained and full data-set}
    \LineComment{$\mathbb{X}$: Set of possible cloud and hyper-parameters configurations}
    \LineComment{$\textbf{Q}$: Set of QoS constraints}
    \vspace{4pt}
    
    \LineComment{Initialization phase}
    \State $\mathcal{T}=\{\langle \textbf{x},s_{i}\rangle : \textbf{x}\in\mathbb{X}, i\in[1,k]\}$ \Comment{Set of untested configs.}
    \State Select at random a configuration $\textbf{x}\in \mathbb{X}$
    \For{$i=1,\ldots,k$} \Comment{Test \textbf{x} with $k$ sub-sampling rates}
           \State $\langle a,c,\textbf{q}\rangle \lam$ Train $M$ in configuration $\langle\textbf{x},s_{i}\rangle$
    \State $\mathcal{S}^A \lam \mathcal{S}^A \cup \{\textbf{x},s_{i},a\}$\Comment{Add accuracy of $\langle \textbf{x},s_{i}\rangle$ to $\mathcal{S}^A$}
    \State $\mathcal{S}^{C} \lam \mathcal{S}^{C} \cup \{\textbf{x},s_{i},{c}\}$ \Comment{Add cost of $\langle \textbf{x},s_{i}\rangle$ to $\mathcal{S}^{C}$}
    \State $\mathcal{S}^\textbf{Q} \lam \mathcal{S}^\textbf{Q} \cup \{\textbf{x},s_{i},\textbf{q}\}$ \Comment{Add QoS constr. of $\langle \textbf{x},s_{i}\rangle$ to $\mathcal{S}^\textbf{Q}$}
    \State $\mathcal{T} \lam \mathcal{T} \setminus \langle\textbf{x},s_{i}\rangle$\Comment{Remove $\langle \textbf{x},s_{i}\rangle$ from  untested configs}
    \EndFor
        \State Fit models $A(\textbf{x},s)$,$C(\textbf{x},s)$,$\textbf{Q}(\textbf{x},s)$ using $\mathcal{S}^A$,$\mathcal{S}^{C}$,$\mathcal{S}^\textbf{Q}$, resp. 

    \vspace{4pt}
    \LineComment{Main optimization loop}
   \For{$(i=1,\ldots,MaxIterations)$}
    \LineComment{Select the most promising configurations using CEA}
    \State $\mathcal{D} \lam$  \{ $\beta$ configs $\langle x,s\rangle\in \mathcal{T}$ with max.~CEA \} \label{line:CEA}
    \LineComment{Acq.~fun.~evaluated only on the configs selected by  CEA}
    \State $\langle \textbf{x}',s'\rangle \lam argmax_{\langle \textbf{x},s \rangle\in\mathcal{D}}  \alpha_{T}(\textbf{x},s)$\Comment{Eq.~\ref{ES1}}
    \State $\langle a,c,\textbf{q}\rangle \lam$ Train $M$ in configuration $\langle \textbf{x}',s'\rangle$
    \State $\mathcal{S}^A \lam \mathcal{S}^A \cup \{\textbf{x}',s',a\}$\Comment{Add accuracy of $\langle \textbf{x}',s'\rangle$ to $\mathcal{S}^A$}
    \State $\mathcal{S}^{C} \lam \mathcal{S}^{C} \cup \{\textbf{x}',s',{c}\}$\Comment{Add cost of $\langle \textbf{x}',s'\rangle$ to $\mathcal{S}^{C}$}
        \State $\mathcal{S}^\textbf{Q} \lam \mathcal{S}^\textbf{Q} \cup \{\textbf{x}',s',\textbf{q}\}$ \Comment{Add QoS of $\langle \textbf{x}',s'\rangle$ to $\mathcal{S}^\textbf{Q}$}
\State $\mathcal{T} \lam \mathcal{T} \setminus \langle \textbf{x}',s'\rangle$\Comment{Remove $\langle \textbf{x}',s'\rangle$ from untested configs}
        \State Fit  $A(\textbf{x},s)$,$C(\textbf{x},s)$,$\textbf{Q}(\textbf{x},s)$ using $\mathcal{S}^A$,$\mathcal{S}^{C}$,$\mathcal{S}^\textbf{Q}$, resp. 
    \State New incumbent $\textbf{x}^{*}$:  feasible config. with max accuracy for $s=1$, as predicted by the models $A(\cdot)$ and $\textbf{Q}(\cdot)$.
    \EndFor
    \EndFunction

    \end{algorithmic}
\end{algorithm}

\section{\ts{}}
\label{sec:trimtuner}

\ts{} is a Bayesian Optimization based approach that jointly optimizes the configuration of the cloud platform over which a ML model is trained as well as the hyper-parameters of the training process.  
More formally, \ts{} considers the following constrained optimization problem:
\begin{equation}
    \label{eq:optProb_Fabulinus}
    \begin{array}[t]{ll}
    \underset{\textbf{x} \in \mathbb{X}}{\text{maximize}} & A(\textbf{x}, s=1) \\
    \text{subject to} & q_{1}(\textbf{x},s=1) \geq 0, \ldots,
    q_{m}(\textbf{x},s=1)  \geq 0
    \end{array}
\end{equation}
\noindent where $\textbf{x}\in \mathbb{X}$    denotes the vector encoding both the cloud's and hyper-parameters' configuration, $s\in [0,1]$ is the sub-sampling rate (relative to the full data-set), $A$ is the accuracy  and $q_{1}\ldots q_{m}$ is a set of Quality of Service (QoS) constraints on the target model (e.g., on the maximum duration/cost of its training process or on the latency for querying the resulting ML model). We assume that the accuracy and constraint functions are unknown, independent, and can only be observed via  point-wise and noisy evaluations by training the ML model in the corresponding configuration. 

Note that both the objective  function (i.e., model's accuracy) and the constraints are expressed for configurations using the full data-set ($s$=1). However,  in order to enhance the efficiency of the optimization process, \ts{}  tests configurations using sub-sampled data-sets ($s<1$) and leverages the knowledge gained via these cheaper evaluations to recommend configurations that use the full data-set.

Algorithm~\ref{alg:trimtuner} provides the pseudo-code of \ts{}. \ts{} receives as input: (i) the ML model ($M$) whose training process has to be optimized along with its (full) training set; (ii)  the set $\mathbb{X}$ of possible cloud and hyper-parameter configurations; (iii) a set of QoS constraints ($\textbf{Q}$).

\mynewpar{Initialization phase.}
As in typical BO approaches, since no \textit{a priori} knowledge is assumed on the target job, \ts{} bootstraps its knowledge base via a random sampling strategy. More precisely, \ts{}   randomly selects a configuration\footnote{Testing more than a configuration provided no benefit in all the tests we performed, but \ts{} supports testing a larger number of  initial configurations, selecting them using Latin Hyper-Cube Sampling~\cite{LHS_original}.} $\textbf{x}\in\mathbb{X}$ and tests it (i.e., trains the model)  using different sub-sampling levels $s_{1},\ldots,s_{k}$.

We choose these sub-sampling levels so that they are biased towards configurations with small data-sets (thus reducing the cost/duration of the bootstrapping phase), while still gathering representative information on how variations of the sub-sampling rate $s$ affect the objective and the constraint functions. Specifically, in the MNIST data-set (which will be evaluated in \S~\ref{sec:eval}), we consider $s\in\{1/60, 1/10, 1/4, 1/2$\}. Note that since we test the same cloud/hyper-parameter configuration and only vary the sampling rate, we can test all the configurations $\langle \textbf{x},s_{i}\rangle$ ($i\in[1,k]$) via a single training instance by taking a snapshot of the accuracy and QoS constraints whenever the sub-sampling rate $s_i$ is achieved --- this yields a cost equivalent to testing a single configuration using 50\% of the model's data-set.

Whenever a configuration $\langle \textbf{x},s \rangle$ is tested, the corresponding accuracy, training cost (due to the use of cloud resources) and QoS constraint values are stored in the $\mathcal{S}^{A}$, $\mathcal{S}^{C}$ and $\mathcal{S}^{\textbf{Q}}$ data-sets, respectively.

The initialization phase ends by building  distinct black-box models (see \S~\ref{subsec:models}) that predict, for an untested configuration $\langle \textbf{x}, s_{i} \rangle$: (i)  the accuracy of the target ML model, noted $A(\textbf{x},s_i)$, using $S^A$ (ii)  the (cloud) cost of its training process, noted $C(\textbf{x},s_i)$, using $S^C$ and (iii) the value of each QoS constraint $q_{1}\ldots q_{m}\in\textbf{Q}$, noted $\textbf{Q}(\textbf{x},s_i)$, using  $S^\textbf{Q}$.


\mynewpar{Main optimization loop.} The optimization loop of \ts{} 
relies  on a novel acquisition function (Eq.~\eqref{ES1}) to determine which configuration to test next:

\vspace{-4mm}
\begin{equation}
\begin{aligned}
    \alpha_{T}(\textbf{x},s)=&
    \mathbb{E}_{p(\textbf{q},a|\textbf{x},s,\mathcal{S})}\!\left[
    \displaystyle\prod^{q_i\in \textbf{Q}} p(q_i(\textbf{x}^*,s\shorteq
    1)\!\geq\!0 | \mathcal{S} \cup
   \{\textbf{x},s,\textbf{q},a\}\right] 
     \\ &\frac{1}{C(\textbf{x},s) }\cdot\mathbb{E}_{p(a|\textbf{x},s,\mathcal{S})} \left[ \int p_{opt}^{s=1}(\textbf{x'}|\mathcal{S}^{A} \cup \{\textbf{x},s,a\})  \right. \\ 
    &  \left. \log \frac{p_{opt}^{s=1}(\textbf{x'}|\mathcal{S}^{A} \cup \{\textbf{x},s,a\}) }{u(\textbf{x'})} d\textbf{x'} \right] 
    \label{ES1}
\end{aligned}
\end{equation}

Conceptually, this acquisition function extends the one of FABOLAS (Eq.~\eqref{AF-F}) by additionally accounting for the probability that the new incumbent, \textbf{x}$^*$, predicted by the models after having acquired knowledge on configuration $\langle \textbf{x},s \rangle$ 
will meet the QoS constraints. The key challenge here is that the new incumbent that the models will predict  after testing $\langle \textbf{x},s \rangle$ is unknown at this point, since we are still  reasoning  on whether to test  $\langle \textbf{x},s \rangle$ or not.

We tackle this problem via a  simulation approach that exploits the current model's knowledge. Intuitively, when computing the acquisition function for $\langle \textbf{x},s \rangle$, the following steps are executed for \textit{all} possible values of accuracy and QoS constraints, $\langle a, \textbf{q}\rangle$:
\begin{enumerate}
\item Extend the accuracy and constraints data-sets ($\mathcal{S}^A$ and $\mathcal{S}^\textbf{Q}$) with $\{\textbf{x},s,a\}$ and $\{\textbf{x},s,\textbf{q}\}$, respectively.
\item Train the models with the extended data-sets.
\item Identify the new incumbent $\textbf{x}^*$ predicted using the updated models. This is the configuration $\langle \textbf{x},s=1\rangle$ that, based on the predictions of the updated accuracy and constraint models,  achieves the largest accuracy among the ones that comply with all constraints.
\item Compute the probability that all constraints are met by the incumbent determined in the previous step, as the product of the  probabilities (predicted by the updated models) that each constraint $q_{i}\in \textbf{Q}$ is respected (recall that constraints are assumed independent).
\end{enumerate}
Finally, the expectation over all possible values of accuracy and QoS constraints $\langle a, \textbf{q}\rangle$ has to be computed. To this end,  the models (prior to being updated) can be used to predict the probability of configuration $\langle \textbf{x},s \rangle$ yielding $\langle a, \textbf{q}\rangle$ for its accuracy and constraints, respectively. The above expectation can be numerically approximated, e.g., using the Gauss-Hermite quadrature~\cite{GH_quadrature}, which, roughly speaking,  approximates the unbounded integral associated with the expectation (that should  be computed over all possible values $\langle a,\textbf{q}   \rangle$) by discretizing the $\langle a,\textbf{q} \rangle$ space over a small number of pre-determined 
``root'' points. To limit the computational complexity of computing $\alpha_T$, in \ts{} we use a  coarser, but cheaper, approximation, which conceptually coincides with using a singe root in the GH quadrature: we simulate the testing of $\langle \textbf{x},s \rangle$ by computing the above steps considering only for $\langle a, \textbf{q} \rangle$ the accuracy and QoS constraints values predicted by the corresponding models in $\langle \textbf{x},s \rangle$.


Note that $\alpha_T$ (see Eq.~\eqref{ES1}) extends the acquisition function of FABOLAS (see Eq.~\eqref{AF-F}) in a modular  way, in the sense that the second and third lines of Eq.~\eqref{ES1}, which correspond to FABOLAS' acquistion function, can be numerically computed as discussed in FABOLAS' paper.

Note also that, for efficiency reasons, \ts{} (see Alg.~\ref{alg:trimtuner} line~\ref{line:CEA}) does not evaluate the acquisition function on every  untested configuration $\langle \textbf{x},s \rangle$, but only 
on a smaller sub-set (denoted $\mathcal{D}$ in the 
pseudo-code) that is determined via a novel filtering 
heuristic, which we called Constrained Expected Accuracy (CEA) and  that we describe in \S~\ref{subsec:CEA}.

Once having determined which configuration $\langle \textbf{x}',s' \rangle\in \mathcal{D}$ maximizes the acquisition function $\alpha_T$, that configuration is tested (by training $M$) and the observed accuracy, cost and constraints' values are stored in the corresponding data-sets. Next, the  models are updated to incorporate the knowledge obtained by testing $\langle \textbf{x}',s'\rangle$ and the new incumbent is recommended.  As already mentioned, the incumbent is selected using the accuracy and constraint models as the configuration with full data-set ($s$=1) that is predicted to be feasible (i.e., meet the constraints with high probability\footnote{We set this probability to 90\% in \ts{}.})  and that achieves maximum accuracy.

\ts{} adopts a simple stop condition that terminates the optimization after a fixed number of cycles. However, it would be relatively straightforward to incorporate more sophisticated, adaptive  stop-conditions~\cite{boTutorial,proteustm} that, e.g., interrupt the optimization if the new predicted incumbent does not improve significantly over the best known optimum.

\subsection{Models} 
\label{subsec:models}

To  estimate the probability distribution of accuracy ($A$), cost ($C$) and constraints ($\textbf{Q}$) for unexplored configurations $\langle \textbf{x},s \rangle$, \ts{} relies on two black-box modelling techniques, namely  Gaussian Processes and ensembles of Decision Trees.

\mynewpar{GPs.} GPs~\cite{gaussianprocesses} represent the \textit{de facto} standard modelling approach in BO, due to their analytical tractability and flexibility~\cite{fabolas,boTutorial}. Key to the tractability of GPs is that the prediction  that they generate follows, by construction, a Gaussian Distribution with known parameters. It is the possibility to define specialized kernels that provides flexibility to GPs, by allowing to imbue the model with domain-specific knowledge.

\ts{}, analogously to FABOLAS, relies on kernels designed to capture the expected impact on cost and accuracy deriving from the use of sub-sampling. Specifically, we use a kernel obtained by the inner product of a ``general purpose" \textit{Matérn} 5/2 kernel and two custom kernels that encode, respectively, the expectation that accuracy and cost of a ML model grow normally with larger data-set sizes (see~\cite{fabolas} for details).

Training GPs is notoriously an expensive  process~\cite{boTutorial}. 
Evaluating the acquisition function of Eq.~\eqref{ES1} (and Eq.~\eqref{AF-F}), requires re-training the models several times. Thus, as we shall see in \S~\ref{sec:eval}, the recommendation process can be extremely slow if a computationally expensive modelling technique, such as GPs, is used.
This led us to explore an alternative approach based on DTs (more precisely Extra Trees~\cite{extraTree}).

\mynewpar{Ensemble of DTs.} DTs are known for their high efficiency, but they  can not be directly used to replace GPs since, unlike  GPs, DTs do not provide a measure of uncertainty of their prediction. We circumvent this problem  by  using an ensemble of DTs and injecting diversity among the various  learners by generating their data-sets drawing with replacement from the same data-set. We then estimate the probability distribution for a prediction as a Gaussian with mean and standard deviation derived from the predictions of the ensemble.

\subsection{CEA}
\label{subsec:CEA}

As mentioned, the numerical computation of \ts{}'s acquisition function (and in general of ES-based acquisition functions) is very expensive, especially if GP models are used. The problem is further exacerbated since \ts{}, differently  from BO-based optimizers like FABOLAS, considers a configuration space that encompasses not only the model's hyper-parameters
, but also the cloud configuration. This results in an exponential growth of the search space and, as such, of the set of untested configurations for which the acquisition function should be evaluated.

The common approach, in the BO literature, to cope with this computational challenge, is to rely on generic search heuristics. These heuristics range from simple random sampling to sophisticated black-box optimizers~\cite{cmaes,direct} and aim at reducing the number of configurations for which the acquisition function is evaluated.
\ts{} instead introduces a novel, domain-specific heuristic called Constrained Expected Accuracy (CEA) (Eq.~\eqref{eq:cea}):
\begin{equation}
\label{eq:cea}
   CEA(\textbf{x},s) = A(\textbf{x},s) \cdot 
   \displaystyle\prod^{q_i\in \textbf{Q}} p(q_i(\textbf{x},s)\geq 0 | \mathcal{S})
  \end{equation}
Defined as the product of the predicted accuracy for a (possibly sub-sampled) configuration  $\langle \textbf{x},s \rangle$  by the  probability that $\langle \textbf{x},s \rangle$ satisfies  the  constraints, CEA  can be seen as a rough, but cheap, approximation of $\alpha_T$. In fact, while CEA directly estimates the quality of a (possibly sub-sampled) configuration $\langle \textbf{x},s \rangle$, $\alpha_T$ predicts how much information the test of  $\langle \textbf{x},s \rangle$ will disclose about $\langle \textbf{x}^{*},s=1 \rangle$, namely the optimal feasible configuration using the full data-set.


Due to its simplicity, CEA can be efficiently evaluated for every untested configuration, and \ts{} uses it to rank and filter configurations:
only the  $\beta\in [0,1]$ configurations with largest CEA are evaluated using $\alpha_T$. We evaluate the sensitivity of \ts{} to the $\beta$ parameter in \S~\ref{subsec:filterheuristic}.

\section{Evaluation}
\label{sec:eval}

\newcommand{\AccuracyC}[1]{\textit{$Accuracy_C$}}
\newcommand{\AccC}[1]{\AccuracyC{}}

This section aims at addressing two main  questions:
\begin{itemize}
    \item Which cost and time savings does \ts{} achieve vs existing BO-based approaches (\S~\ref{subsec:imprs})?
    \item How expensive is the recommendation process in \ts{} and how effective is the CEA heuristic in accelerating it (\S~\ref{subsec:filterheuristic})?
\end{itemize}



\mynewpar{Data-sets.}
The data-sets used to evaluate \ts{} were obtained by training three different neural networks (NN) in the AWS cloud: Convolutional Neural Network (CNN), Multilayer Perceptron (MLP), and Recurrent Neural Network (RNN). The networks were implemented using the Tensorflow framework~\cite{tensorflow} and trained on the MNIST database~\cite{mnist}. 

\begin{table}[t]
\vspace{-4pt}\footnotesize
\caption{TensorFlow (top) and cloud (bottom) parameters.}
\label{tab:hyperandcloud}
\centering
\begin{center}
\begin{tabular}{ll}
\toprule
\textbf{Parameter}  & \textbf{Values}                   \\ \midrule
Learning rate       & $\{10^{-3}, 10^{-4}, 10^{-5}\}$   \\
Batch size          & $\{16, 256\}$                     \\ 
Training mode       & \{sync, async\}                   \\ 
Data-set size [\%]  & $\{1.67, 10, 25, 50, 100\}$       \\
\bottomrule
\vspace{1pt}
\end{tabular}

\begin{tabular}{lll}
\toprule
\textbf{VM type} & \textbf{VM characteristics} & \textbf{\#VMs} \\
\midrule
t2.small    & \{1 VCPU, 2 GB RAM\}  & \{8, 16, 32, 48, 64, 80\} \\ 
t2.medium   & \{2 VCPU, 4 GB RAM\}  & \{4, 8, 16, 24, 32, 40\}  \\
t2.xlarge   & \{4 VCPU, 16 GB RAM\} & \{2, 4, 8, 12, 16, 20\}   \\
t2.2xlarge  & \{8 VCPU, 32 GB RAM\} & \{1, 2, 4, 6, 8, 10\}     \\ \bottomrule
\end{tabular}
\end{center}
\end{table}

Each configuration (see Table~\ref{tab:hyperandcloud}) is composed of cloud resources (number, type, and size of the virtual machines), application-specific parameters (batch size, learning rate, and training mode), and the size of the data-set used to train the NN. This results in a search space of 1440 configurations  (288 configurations for each data-set size). To reduce noise in the measurements, we executed the training of the NN  in each configuration three times.
Whenever we train a model in a configuration $\langle \textbf{x},s\rangle$ we measure the achieved accuracy, training time and cloud cost. Collecting these data-sets took more than 2 months and costed a total of about 1200 USD. We have made these data-sets publicly available\footnote{\url{https://github.com/pedrogbmendes/TrimTuner.git}}, along with the implementations of \ts{} and of the considered baselines.

Based on  these data-sets, we define a QoS constraint  that  limits the maximum training cost to be $\$0.02$, $\$0.06$, and $\$0.1$ for RNN, MLP, and CNN, respectively. Table~\ref{tab:feasible_space_fabulinus} reports the number (percentage) of configurations that use the full data-set ($s$=1) and  comply with the cost constraints, and the number of those whose accuracy is no more than 5\% lower than the accuracy of the feasible configuration with highest accuracy. Only around 10\% of the full data-set configurations are close to the optimum,  which illustrates the non-triviality of the considered optimization problem.

 \begin{table}
\centering
\footnotesize
\caption{Number of feasible configurations with an accuracy within $5\%$ of the accuracy of the feasible configuration with highest accuracy for the full data-set.}
\label{tab:feasible_space_fabulinus}
\begin{tabular}{lcc}
\toprule
\textbf{Neural} & \textbf{Feasible} & \textbf{Feasible Configurations }\\
\textbf{Network} & \textbf{Configurations} & \textbf{with high accuracy}  \\
\midrule
RNN  & 178 (61.8\%) & 28 (9.72\%) \\
MLP & 161 (55.8\%) &  29 (10.07\%) \\
CNN & 111 (38.5\%) & 39 (13.54\%) \\ 
\bottomrule
\end{tabular}
\end{table}

\mynewpar{Baselines.}
We compare \ts{} using GPs and DTs  against constrained Expected Improvement (EIc) and EIc/USD, two popular acquisition functions for constrained optimization problems that were used by two recent cloud  optimizers (CherryPick~\cite{cherrypick} and Lynceus~\cite{lynceus}, resp.).

 EIc extends EI (Eq.~\eqref{eq:ei}) by factoring in the probability that the configuration being evaluated will meet the constraints. EIc/USD, in its turn, extends EIc by considering the trade off between the benefits stemming from an exploration (computed using EIc) and the exploration cost (estimated via a dedicated model). None of these techniques use sub-sampling. We use GPs as base models for both EIc and EIc/USD that were implemented using the George library~\cite{george} in Python.
 We include in the comparison also Fabolas~\cite{fabolas}, which uses sub-sampling but does not consider constraints, and a simple random approach. We used the publicly available standard implementation of FABOLAS.

\mynewpar{Experimental setup and evaluation.}
The reported results are the average of  10 independent runs.
%
%
%
%
%
%
For all the compared  systems we bootstrap the models using 4 initial samples. For \ts{} and FABOLAS, which use sub-sampling strategies, we select a cloud and hyper-parameter configuration uniformly at random, and test it over the considered 4 data-set sizes (Table~\ref{tab:hyperandcloud}). For EIc and EIc/USD, which do not use sub-sampling, we sample 4 full data-set configurations using  Latin Hypercube Sampling (LHS).

We set the maximum number of iterations to 44 for all optimizers.
Unless otherwise stated, for both \ts{} variants we use the CEA heuristic, setting the filtering rate $\beta$ to 10\%.

All the systems were implemented in Python3.6 and the simulations were deployed in a VM running Ubuntu 18.04 LTS with 32 cores and 8GB of memory, hosted in machines with a Intel Xeon Gold 6138 CPU and 64GB of memory.

\mynewpar{Evaluation Metrics.}
To evaluate the systems, we use a metric which we named \textit{Constrained Accuracy} (\AccC{}), that penalizes recommended configurations that do not respect the cost constraint. 

\vspace{-3mm}
\begin{equation}
    \label{eq:accuracyc}
    \AccC{}=  \begin{cases} A(x,s), & \mbox{if } (x,s)\mbox{ is feasible} \\
    A(x,s) \cdot \frac{C_{max}}{C(x,s)}, & \mbox{otherwise } \end{cases}
\end{equation}
It is easy to  see that this metric imposes larger penalizations to configurations that violate the constraint by a larger extent. 


\subsection{Comparison with state of the art optimizers}
\label{subsec:imprs}

\begin{figure*}[t]
  \begin{subfigure}[h]{0.33\textwidth}
        \includegraphics[scale = 0.55]{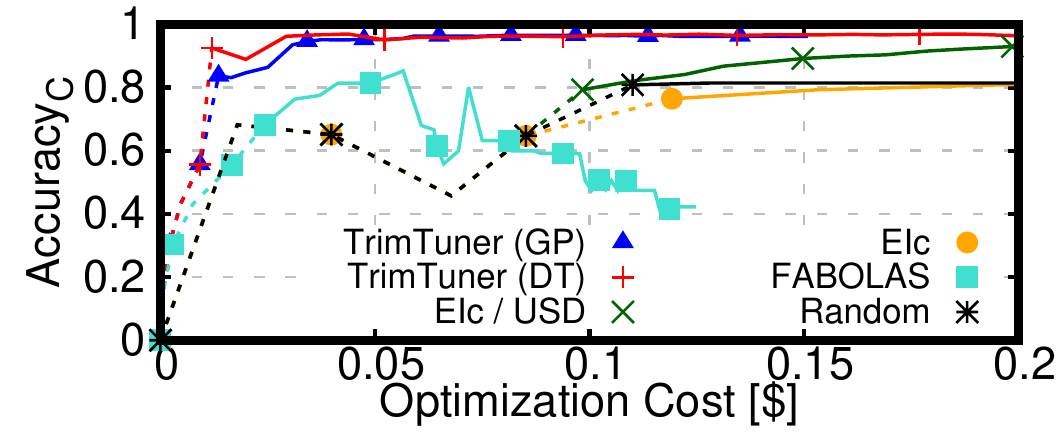}
        \caption{RNN}
        \label{fig:optimizers_cost_rnn}
    \end{subfigure}
    \begin{subfigure}[h]{0.33\textwidth}
        \includegraphics[scale = 0.55]{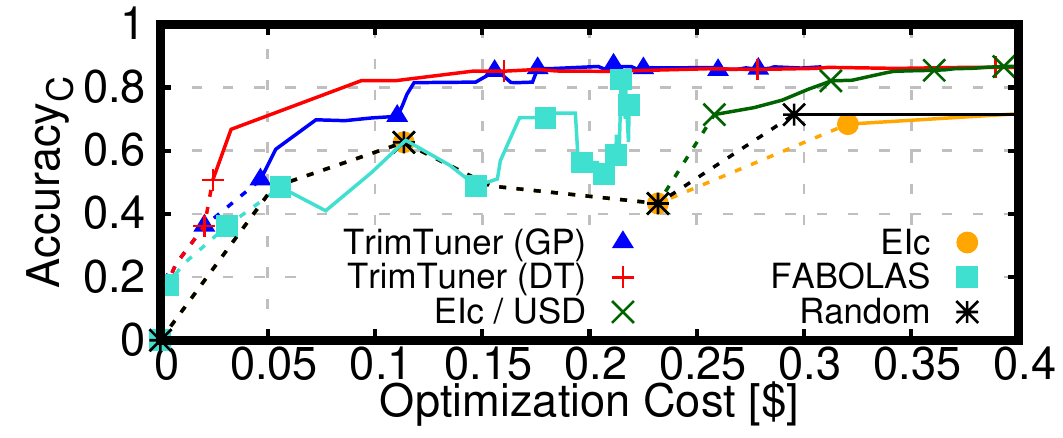}
        \caption{MLP}
        \label{fig:optimizers_cost_ml}
    \end{subfigure}
    \begin{subfigure}[h]{0.33\textwidth}
        \includegraphics[scale = 0.55]{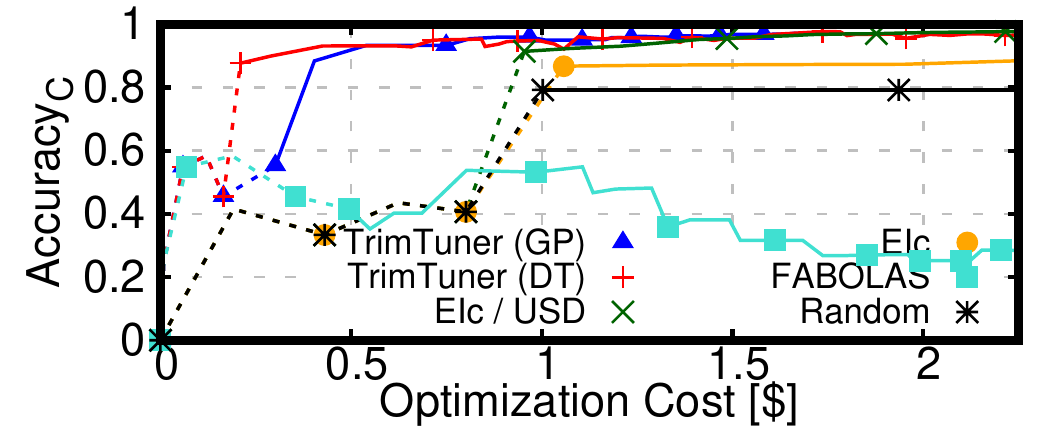}
        \caption{CNN}
        \label{fig:optimizers_cost_cnn}
    \end{subfigure}
    \vspace{-2.5mm}
\caption{\AccC{} for RNN, MLP, and CNN  as a function of the optimization cost. The dashed lines correspond to the initialization phase, during which a random sampling policy is used by all solutions.}
\label{fig:opt_cost}
\end{figure*}

Figure~\ref{fig:opt_cost} evaluates the cost efficiency of the compared solutions by reporting the  \AccC{} of the  recommended incumbent as a function of the cost of the optimization process for the various networks. The plots clearly show that both \ts{} variants achieve higher \AccC{} levels at a fraction of the cost of the other solutions. The reason underlying \ts{}'s gains  with respect to EIc and EIc/USD is the use of sub-sampling, whose benefits are clear both in the initialization stage (shown using dashed lines) and once the models are in use. The average cost of each exploration step with \ts{} is approx. 10$\times$ and 2.4$\times$ smaller than with EIc and EIc/USD, respectively. This is explicable by considering that \ts{} uses an average sub-sampling rate of approx. 1/4 whereas EIc and EIc/USD test using full data-sets.

As expected, FABOLAS is the worst performing solution in this constrained optimization problem (for which it is not designed): since FABOLAS does not keep into account constraints, although the configurations it recommended achieve high accuracy, they frequently violate the cost constraint.

As for the cost efficiency of the optimization process for the two variants of \ts{}, the plots do  not highlight significant differences. DTs appear to be slightly more accurate than GPs in the considered data-sets, confirming that they can represent a solid (and as we will see shortly way more lightweight) modelling alternative to GPs.

\begin{figure}
    \begin{subfigure}[h]{0.22\textwidth}
        \includegraphics[scale = 0.6]{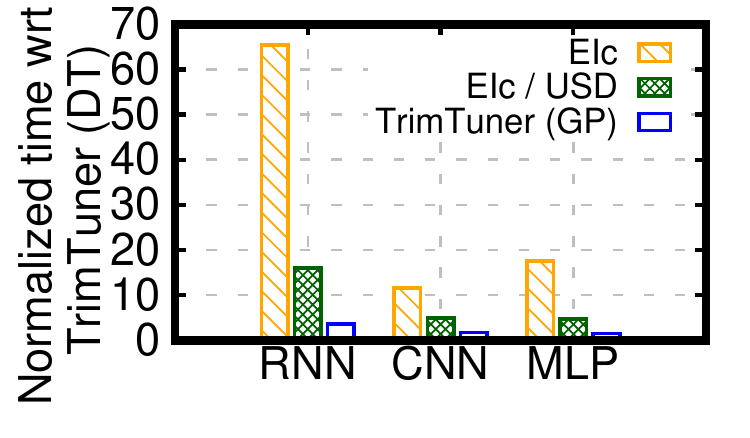}
        \caption{Time speed ups.}
        \label{fig:speed_ups}
    \end{subfigure}
    \hspace{4pt}
    \begin{subfigure}[h]{0.22\textwidth}
        \includegraphics[scale = 0.6]{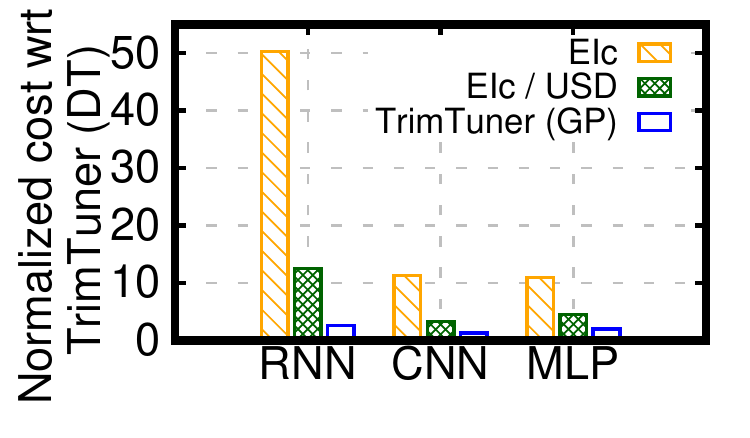}
        \caption{Cost savings.}
        \label{fig:cost_savings}
    \end{subfigure}
    \caption{Time (\ref{fig:speed_ups}) and cost (\ref{fig:cost_savings}) savings achieved by \ts{} when using an ensemble of decision trees.}
    \label{fig:trimtuner}
\end{figure}

Figure~\ref{fig:trimtuner} provides another perspective to assess the gains achieved by \ts{} w.r.t. the considered baselines by reporting the time (Figure~\ref{fig:speed_ups}) and cost (Figure~\ref{fig:cost_savings}) savings that \ts{} (DT  variant) achieves to identify a configuration whose accuracy is  90\% (or more) of the optimal (feasible) solution. We omit FABOLAS and Random from the  plots, which perform poorly, to enhance visualization. Using EIc and EIc/USD, the optimization process lasts up to 65$\times$/15$\times$ (resp.) and costs up to 50$\times$/10$\times$ more.





Table~\ref{tab:timesRecommendopt} reports the average time to recommend the next configuration to test, allowing us  to compare the computational complexity  of the considered solutions. As expected, EIc and EIc/USD, which use the simplest acquisition functions, although with GPs, have the lowest computational cost. FABOLAS, which also relies on GP-based models, takes approximately one order of magnitude longer to output a recommendation ($\approx$14 minutes) and the GP-based variant of \ts{} takes approximately 30\% longer. It is interesting to see that the DT-based implementation of \ts{} achieves a 13$\times$ speed up compared to the GP variant, attaining a performance almost on par with EIc and EIc/USD.




To sum up, \ts{} can recommend configurations that achieve high accuracy and meet the cost constraint, while obtaining  significant cost and time reductions w.r.t. the considered baselines through the use of sub-sampling.

\begin{table}[t]
\centering
\footnotesize
\caption{Average time to recommend a configuration (average of the three data-sets).}
\label{tab:timesRecommendopt}
\begin{tabular}{lSS}
\toprule
\textbf{Optimizer} & \textbf{Avg. time to recommend} & \textbf{Standard}\\
& \textbf{ a configuration [min]} & \textbf{Deviation} \\ \midrule
\ts{} (GPs) & 18.65 & 2.31    \\ 
\ts{} (DTs) & 1.36 & 0.28    \\ 
Fabolas   & 13.96  & 1.88    \\ 
EIc (or EIc/USD) & 1.17  & 0.07      \\
\bottomrule
\end{tabular}

\end{table}

\subsection{Sensitivity to Filtering Heuristic}
\label{subsec:filterheuristic}

The data reported in Table~\ref{tab:timesRecommendopt} was obtained by using in both \ts{} variants the CEA heuristic configured to select   $10\%$ ($\beta$) of the untested configurations. This section investigates the efficiency of this heuristic and the sensitivity of \ts{} to the setting of the $\beta$ parameter.

\begin{figure}[t]
    \centering
    \includegraphics[scale = 0.6]{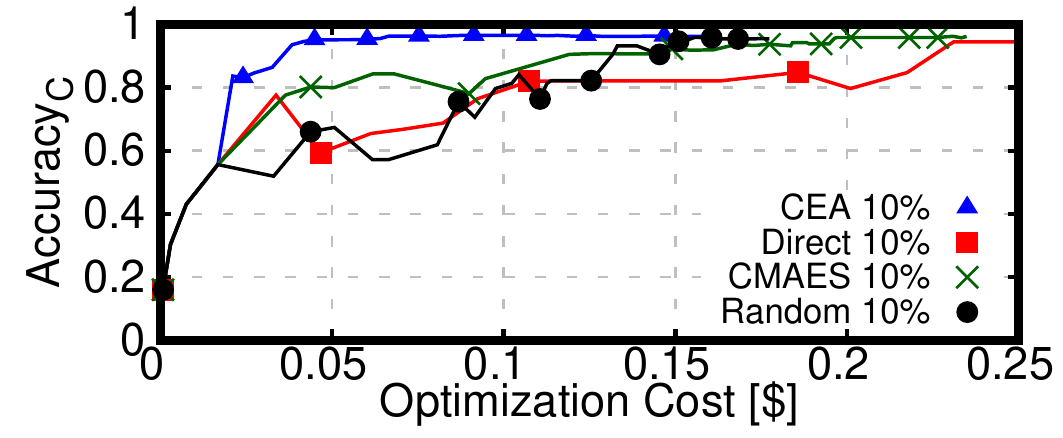}
    \vspace{-2mm}
    \caption{Comparison of the optimization cost for RNN using \ts{} (GP variant) using different filtering heuristics.}
    \label{fig:trimtuner_filterHeuristic}
\end{figure}

We start by comparing, in Figure~\ref{fig:trimtuner_filterHeuristic},  the \AccC{} that is achieved in RNN when using \ts{} with GPs and the following heuristics: CEA,  two state of the art black-box optimizers (used, e.g., in FABOLAS), namely Direct~\cite{direct} and CMAES~\cite{cmaes} and a simple random policy. For all the heuristics we set the filtering level ($\beta$) to 10\% and treat the optimization cost as the independent variable.

The plot confirms the cost-efficiency of the proposed heuristic: considering the cost spent to recommend a feasible configuration whose  accuracy is 90\% (or more) of the optimum, CEA achieves a 3.62$\times$ and a 7$\times$ savings when compared to CMAES and Direct, respectively.



Table~\ref{tab:timesRecommend} allows to evaluate the computational efficiency of CEA by comparing the average time to recommend  the next configuration using \ts{} (both variants) with different  heuristics and filtering levels (including no filter).  We start by observing that when considering a 10\% filtering level, recommending configurations with CEA takes roughly as long as with Random. Note that with a random policy, the time taken to recommend a  configuration is  dominated by the evaluation of the acquisition function (for the  configurations selected by the filtering heuristic). This data confirms that CEA is, indeed, a lightweight filtering heuristic, that is actually more computationally efficient than Direct and CMAES (by up to approximately $2\times$).

\begin{figure}[t]
    \centering
    \includegraphics[scale = 0.6]{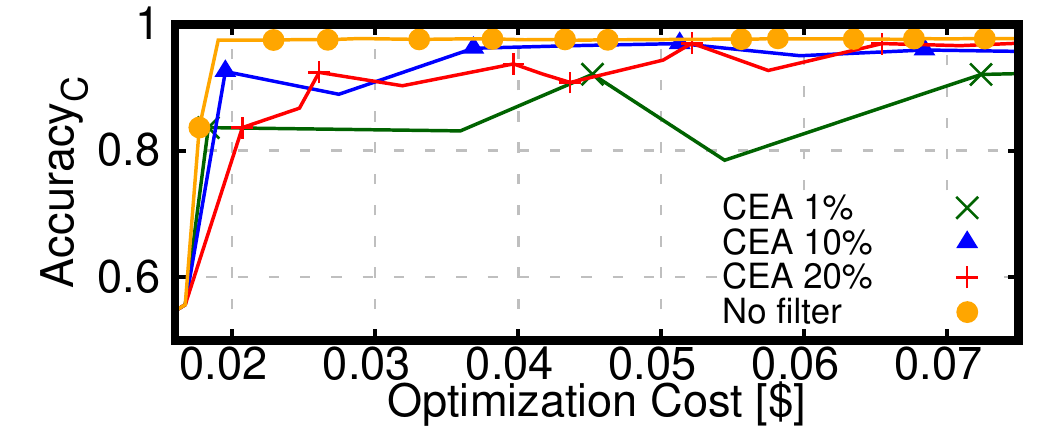}
    \vspace{-3mm}
    \caption{Sensitivity study on the $\beta$ paramater for CEA (RNN,  DT). The initialization phase does not use CEA and is omitted.}
    \label{fig:trimtuner_cea}
\end{figure}

Finally, in  Figure~\ref{fig:trimtuner_cea}, we report the results of a sensitivity study on the  impact of tuning the filtering level ($\beta$) with CEA. As expected, the best results are achieved when the no filtering heuristic is employed. However, this comes at a very high computational cost (see Table \ref{tab:timesRecommend}): if no filtering heuristic is used, it takes on average about two hours to recommend a configuration for the GP variant of \ts{}; with DTs, that time reduces to approximately 4 minutes, but still remains quite large. Overall, these experimental data confirm the relevance of developing effective filtering heuristics.

As for the sensitivity to the tuning of $\beta$, clearly the smaller the number of configurations that the heuristic can select, the worse the performance. Yet, we do not observe a significant degradation for values of $\beta$ as low as 10\%, which motivates the setting employed in the study presented in Section~\ref{subsec:imprs}.

\begin{table}[t]
\footnotesize
\caption{   Average time to recommend the next configuration with different heuristics and filtering levels (RNN, \ts{}).}
\label{tab:timesRecommend}
\centering
\begin{tabular}{lSS}
\toprule
\textbf{Filtering} & \textbf{Recommendation time} & \textbf{Recommendation time} \\
\textbf{Heuristic} & \textbf{\ts{} (GPs) [min]} & \textbf{\ts{} (DTs) [min]} \\
\midrule
No filter      & 125.76  & 3.69 \\
CEA (1\%)      & 5.94   & 1.07  \\ 
CEA (10\%)     & 16.85  & 1.72 \\ 
CEA (20\%)     & 28.65  & 2.05 \\ 
Direct (10\%)  & 36.18  & 2.63  \\ 
CMAES (10\%)   & 30.87  & 2.26  \\
Random (10\%)  & 16.53  & 1.62 \\ 

\bottomrule
\end{tabular}
\end{table}

\section{Conclusion}
\label{sec:conclusion}

We presented \ts{}, a  system for optimizing the training of ML jobs in the cloud that exploits data sub-sampling  techniques to enhance the efficiency of the optimization process. \ts{} builds upon  recent systems for hyper-parameter tuning, which it extends by supporting user-defined constraints, and jointly optimizing cloud and model's hyper-parameters. \ts{} relies on two methods to accelerate the recommendation process: a new heuristic, called Constrained Expected Accuracy; and the adoption of an ensemble of decision trees to model the cost and accuracy.

Thanks to sub-sampling,  \ts{}  reduces the cost and latency of the exploration process by up to 50$\times$ and 65$\times$, 
respectively, whereas the  joint use of CEA and decision trees accelerates the recommendation process by up to 117$\times$.

In future work, we plan to extend \ts{} to cope with alternative optimization problems, e.g.,  multi-objective optimization of both cost and accuracy, and to evaluate it in problems with multiple constraints.

\bibliographystyle{IEEEtran}
\bibliography{biblio}

\end{document}